\documentclass[10pt,twocolumn,letterpaper]{article}

\usepackage{cvpr}
\usepackage{times}
\usepackage{epsfig}
\usepackage{graphicx}
\usepackage{amsmath}
\usepackage{amssymb}
\usepackage{mathrsfs}
\usepackage{booktabs}
\usepackage{multirow}
\usepackage{xspace}
\usepackage{subcaption} 
\newcommand{\sysname}{TAT\xspace}
\newcommand{\argmin}{\operatornamewithlimits{argmin}}

\newcommand{\vect}[1]{\boldsymbol{#1}}

\newcommand{\para}[1]{\smallskip\noindent\textbf{#1}}

\usepackage[breaklinks=true,bookmarks=false]{hyperref}

\cvprfinalcopy 

\def\cvprPaperID{****} 

\begin{document}

\title{Tracklet Association Tracker: \\
An End-to-End Learning-based Association Approach for Multi-Object Tracking}

\author{
Han Shen\\
IIIS, Tsinghua Univeristy\\
Beijing, China\\
{\tt\small shenh15@mails.tsinghua.edu.cn}
\and
Lichao Huang\\
Horizon Robotics\\
Beijing, China\\
{\tt\small lichao.huang@hobot.cc}
\and
Chang Huang\\
Horizon Robotics\\
Beijing, China\\
{\tt\small chang.huang@hobot.cc}
\and
Wei Xu\\
IIIS, Tsinghua Univeristy\\
Beijing, China\\
{\tt\small weixu@mail.tsinghua.edu.cn}
}
\maketitle

\begin{abstract}
Traditional multiple object tracking methods divide the task into two parts: affinity learning and data association. The separation of the task requires to define a hand-crafted training goal in affinity learning stage and a hand-crafted cost function of data association stage, which prevents the tracking goals from learning directly from the feature. In this paper, we present a new multiple object tracking (MOT) framework with data-driven association method, named as Tracklet Association Tracker (\sysname).  The framework aims at gluing feature learning and data association into a unity by a bi-level optimization formulation so that the association results can be directly learned from features. To boost the performance, we also adopt the popular hierarchical association and perform the necessary alignment and selection of raw detection responses. Our model trains over $20\times$ faster than a similar approach, and achieves the state-of-the-art performance on both MOT2016 and MOT2017 benchmarks. 
\end{abstract}

\section{Introduction}
Multiple Object Tracking (MOT) is one of the most critical middle-level computer vision tasks with wide-range applications such as visual surveillance, sports events, and robotics. Owing to the great success of object detection techniques, detection based paradigm dominates the community of MOT.  The critical components of the paradigm include an \emph{affinity model} telling how likely two objects belong to a single identity, and a \emph{data association method} that links objects across frames, based on their affinities, so as to form a complete trajectory for each identity.

Tracklet-based association is a well-accepted approach in detection-based MOT~\cite{huang-et-al:2008,zamir-et-al:2012,wang-et-al:2017,tang-et-al:2015}. It is usually constructed by two stages: In stage I, we link detection responses in the adjacent frame using straightforward strategies to form short tracklets. In stage II, we mainly perform two tasks: extract much finer features from the tracklets, including temporal and spatial, appearance and motion data to construct a tracklet-level affinity model, and then perform graph-based association across all of them, and conduct necessary post-processing. There are two advantages of this approach, compared to associations on detection responses directly. With tracklet-based association, the number of connected components is significantly brought down so that investigating detection dependency across distant frames is computationally affordable. 
Besides, it is capable of extracting high-level information, while reducing bounding box noises brought by bad detectors~\cite{huang-et-al:2008}.

There are various ways to define the affinity model in stage I, like bounding box intersection-over-union(IOU), spatial-temporal distance, appearance similarity, \etc. The harder part exists in stage II. For the affinity model, traditional hand-crafted features or individually learned affinities do not work well~\cite{huang-et-al:2008,zhang-et-al:2008}, due to the lack of data-driven properties in joint consideration of multiple correlated association choices. For the association, it is regular to use a global optimization algorithm, such as linear programming or network flow, to link these short tracklets.  However, it is non-trivial to define a proper cost function for these approaches. 
Earlier trackers use hand-crafted cost functions and perform an inference afterward. Sometimes, they have to use grid search and empirical tuning to find the hyper-parameters producing the best outcome.  
%

Recently, deep learning has shown its powerful learning capability in feature extraction. It outperforms almost all hand-crafted feature descriptors, such as HOG, SIFT, \etc. Large-scale data provides nutrients for the learning of large models, and data-driven approaches are becoming rather important. However, for MOT, the ultimate goal, like MOTA, is not directly related to the features of objects. Thus, it is necessary to model the connectivity across frames into an association method, so that we can build a bridge from the features to the association goal, and perform learning using optimization method. Because network flow is an approach which can be solved in polynomial time, it has a great potential for data-driven learning comparing to other NP-hard formulations~\cite{tang-et-al:2015,tang-et-al:2016,tang-et-al:2017,zamir-et-al:2012}. Schulter \etal~\cite{schulter-et-al:2017} propose a network flow based novel framework to handle both tasks of stage II in an \emph{end-to-end} fashion: by back-propagating a constrained linear programming objective function.  While the framework allows learning from detection features and auto-tuning of costs, the drawbacks are clear: 1) the large number of detection responses limit the expansion of window size; 2) the unbounded costs are easily diverging, and training is slow; 3) High-level motion information is not considered.  


We propose Tracklet Association Tracker (\sysname), an improved bi-level optimization framework compared to work of Schulter \etal~\cite{schulter-et-al:2017} in three key aspects.  First, we use deep metric learning to extract the appearance embedding for each detection response; Second, we introduce tracklet to the framework, not only accelerating the computation but also provides motion dependency.  Last but not the least, we adopt an approximate gradient that significantly improves the association model training process. By clarifying the boundary of cost values, the framework ensures convergence can always be achieved and includes all cost parameters into the end-to-end training process while retaining high accuracy.




All in all, our contributions include:
\begin{itemize}
\item We introduce tracklet association into the bi-level optimization framework. By exploiting tracklets, our system improves the performance on long time occlusion.  

\item We implement \sysname, an approximate network flow learning approach that provides a more stable and faster(over $20\times$) solution of similar method~\cite{schulter-et-al:2017}. The method achieves the state-of-the-art performance on MOT2016 and MOT2017~\cite{milan-et-al2:2016}.

\item We conduct comprehensive discussions on the impact of each component we introduce. Besides, we give a quantitative evaluation on the importance of alignment and noisy outlier removal, which shows both ancient and modern detectors can benefit from these strategies.
\end{itemize}

\section{Related Work}
\label{sec:related_work}

\begin{figure*}[t]
	\begin{center}
	\includegraphics[width=1.\linewidth]{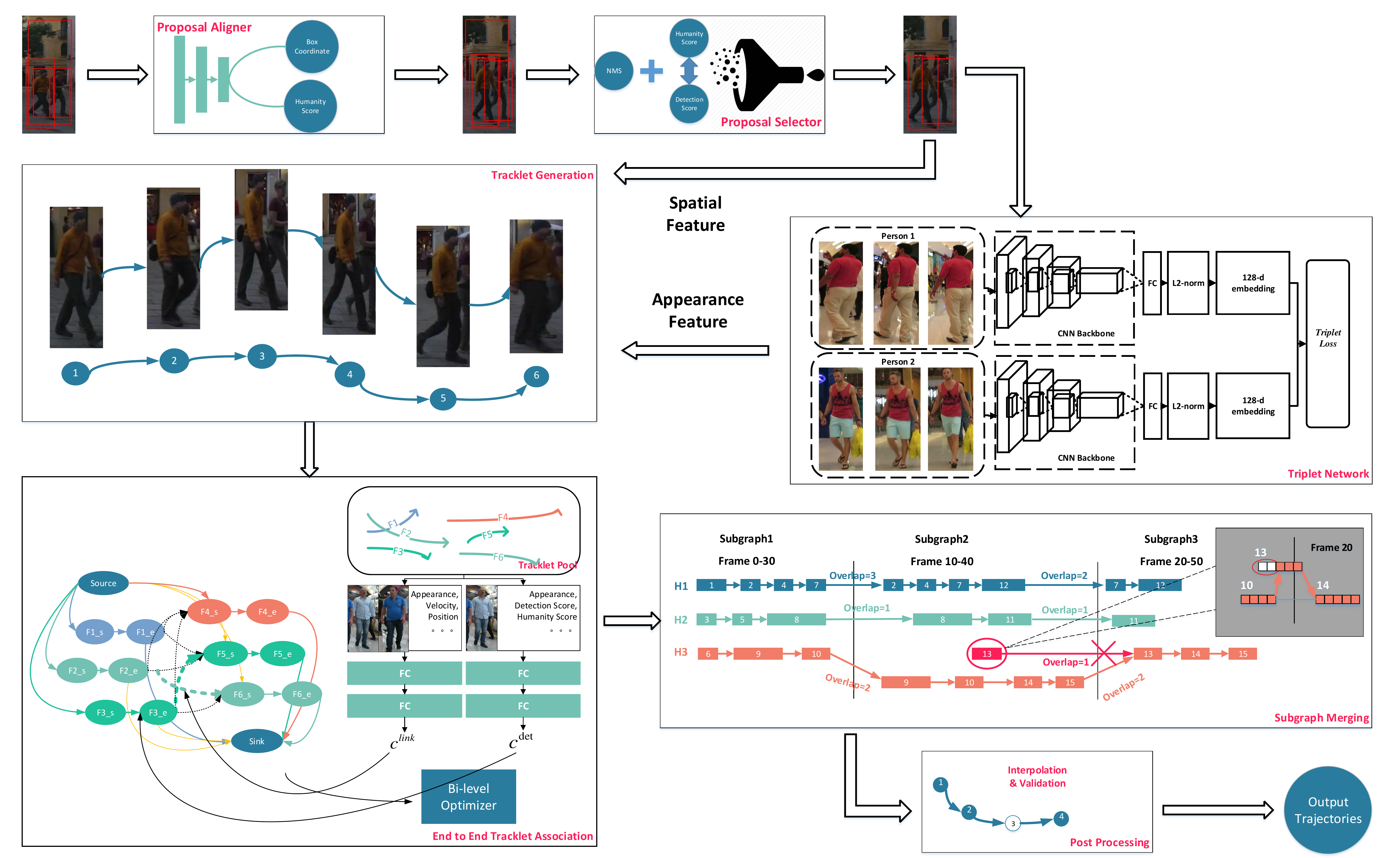}

	\end{center}
		\vspace{-2mm}

	\caption{
%
\sysname workflow. 1) The detection responses are processed by the ``Proposal Aligner'' and ``Proposal Selector'' to produce aligned and high-quality bounding boxes. 2) They are fed into ``Triplet Network'' to get the corresponding appearance embeddings. 3) Then, we use a simple Multilayer Perceptron (MLP) to link neighboring boxes into tracklets according to the appearance and spatial features, a.k.a ``Tracklet Generation''. 4) As the core step of \sysname, we propose an end-to-end network flow association approach (bridged on bi-level optimization) to learn costs $\vect{c}^{det}, \vect{c}^{link}$ from the tracklet feature, a.k.a ``Tracklet Association''. The mappings from $\vect{c}^{det}, \vect{c}^{link}$ to the edges of a network flow graph are shown by the black arrows in the ``End to End Tracklet Association'' module. 5) To reduce graph size, the sequences are divided into subgraphs, so there are strategies to merge results of subgraphs into long trajectories, as shown in the ``Subgraph Merging'' module. 
6) Finally, in the post-processing step, we do interpolation and post-validation to generate the final trajectories.
	}
	\vspace{-2mm}
\label{fig:workflow}
\end{figure*}


Since the tracking-by-detection approach becomes the mainstream in multi-target tracking, data association is regarded as the core part of MOT ~\cite{luo-et-al:2014}. There are a lot of ways to solve data associations.
  Widely adopted probabilistic inference methods like Kalman filter~\cite{rodriguez-et-al:2011}, extended Kalman filter~\cite{mitzel-leibe:2011} and particle filter~\cite{breitenstein-et-al:2009} rely on the first-order Markov assumption to estimate position in the new frame based on previous states and present observations. New detections can be assigned locally between adjacent or nearby frames using bipartite algorithms such as the optimal Hungarian
algorithm~\cite{huang-et-al:2008,xing-et-al:2009}, or k-partite graph matching~\cite{collins:2012}. Multiple Hypothesis Tracking (MHT)~\cite{blackman:2004,kim-et-al:2015} postpones determinating ambiguous matches until enough information obtained. It results in a combinatorially increasing search space so that the hypothesis tree is pruned regularly. These local data association based tracking are sensitive to occlusion and noisy detections.

Tracking algorithms with global or delayed optimization~\cite{blackman:2004,kim-et-al:2015} try to produce longer and more robust trajectories by considering more frames or even the entire sequence in one shot. A popular paradigm is to formulate the task of MOT as an equivalence of solving an extremum of graph~\cite{tang-et-al:2015,tang-et-al:2016,tang-et-al:2017,zamir-et-al:2012,brendel-et-al:2011}. For instance, multi-cut-based approaches\cite{tang-et-al:2015,tang-et-al:2016,tang-et-al:2017} try to decompose graph into isolated components with multi-cut algorithm, so that detections in same component belong to same identity. Generalized Minimum Clique Graph\cite{zamir-et-al:2012} treats detection association as to find the minimum clique in a corresponding graph. However, these graph-based approaches are NP-hard, indicating that only sub-optimal solution can be achieved even with expensive approximate methods.

 Exceptionally, network-flow-based tracking~\cite{zhang-et-al:2008,pirsiavash-et-al:2011,dehghan-et-al2:2015,butt-collins:2013} uses graph formulation and can be solved in polynomial time. It restricts the cost function to contain only unary and pairwise terms to achieve efficient inference. For instance, the work of Zhang \etal~\cite{zhang-et-al:2008} and Pirsiavash \etal~\cite{pirsiavash-et-al:2011} both assume logarithm cost functions and solve min-cost max-flow by push-relabel or successive shortest path algorithms. Dehghan \etal~\cite{dehghan-et-al2:2015} use network flow to simultaneously predict detections and associate identities. Butt and Collins~\cite{butt-collins:2013} encode neighboring connections differently so they leverage network flow to capture the relationships among three consecutive frames. Our work is most similar to the work of Schulter \etal~\cite{schulter-et-al:2017} in that we both formulate the MOT problem in network flow paradigm and solve it by a bi-level optimization problem. Our work differs in the hierarchical design and efficiency.
 

Recently, people use deep learning to improve the tracking performance. DeepMatching~\cite{tang-et-al:2016} applies a convolutional neural network (CNN) to yield non-rigid matching between image pairs. Sadeghian \etal~\cite{sadeghian-et-al:2017} encode history trajectory into embeddings by long short-term memory(LSTM)~\cite{hochreiter-schmidhuber:1997} and compute embedding affinity with present detections.
Quadruplet CNN~\cite{son-et-al:2017} simultaneously train a multi-task loss to jointly learn object appearance and bounding box regression, and adopt minimax label propagation for matching. Instead, our work uses triplet loss and solves the association problem using a learnable association framework.


\section{MOT Framework}
\label{sec:framework}

Fig.~\ref{fig:workflow} illustrates the fundamental steps in \sysname. The components include 1) a proposal aligner, a proposal selector and a triplet network to achieve accurate appearance model; 2) a tracklet generation module to connect neighboring bounding boxes; 3) an end-to-end bi-level optimization tracklet association module to associate tracklets. 4) subgraph merging and post-processing module to propose final trajectories.  In this section, we will elaborate on each step.

\subsection{Appearance Model}
\label{sec:appearance}
\para{Proposal aligner.} To train an appearance model with high discriminative ability, it is essential to have bounding boxes aligned with targets; otherwise, the appearance model is prune to obscurity. Ancient detectors suffers from localization accuracy~\cite{felzenszwalb-et-al:2010}, while modern detectors are limited by the variance of preset anchor sizes and aspect ratios~\cite{ren-et-al:2015}. Thus, a secondary alignment is beneficial for that detections can be treated as better anchors compared to preset ones.
Hence, we adopt a region proposal aligner with convolutional neural network architecture. It takes the slightly padded image patches of the corresponding detection responses as input, the aligned bounding box offset $\Delta \vect{p}$ and their respective classification scores $\vect{h}$ as output. It allows us to treat the raw detections as anchors, and to perform a regression enhancement based on the accurate baseline.


\para{Proposal selector.} After aligning the detection boxes, there are cases where two boxes of a single target overlap larger than before. We use non-maximum-suppression (NMS) to remove duplicates. Further, we use the classification score of the proposal aligner as a new indicator, naming it \emph{humanity} for that it reflects the probability of the target being a human w.r.t. the new box coordinates.  We filter out the boxes with both low humanity and detection score because true positives are unlikely to perform badly simultaneously in two measures, hence we have high confidence to regard the removed boxes as false positives. Through proposal selection, the remaining proposals are cleaner and are less likely to result in redundant overlapping trajectories.

\begin{figure}[t]
\begin{center}
   \includegraphics[width=1.\linewidth]{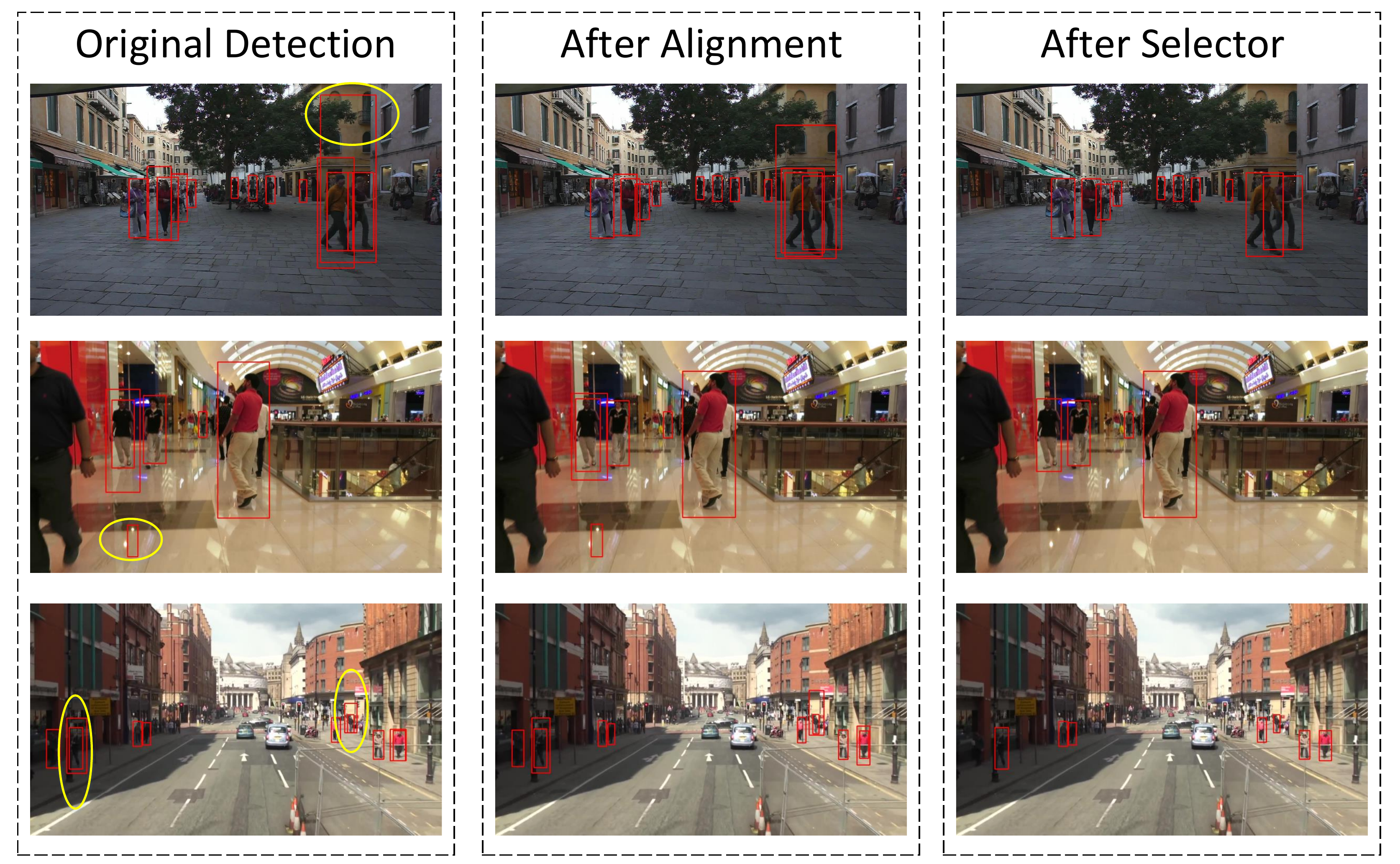}
\end{center}
\vspace{-4mm}
\caption{The proposal aligner and selector demonstration. The yellow ovals indicate false positives of DPM~\cite{felzenszwalb-et-al:2010}.}
   
\label{fig:regression_demo}
\vspace{-2mm}

\end{figure}

\para{Triplet Network.} In a surveillance video, it is common to assume that human appearances do not change much in a short period. Based on the cleaned proposals processed by the proposal aligner and selector, we use metric learning~\cite{schroff-et-al:2015,chen-et-al:2017,leal-et-al:2016,wang-et-al:2017,son-et-al:2017} to learn an embedding $\vect{z}$ for each tracking target candidate so that the distance between two targets with the same identity is smaller than that with different identities. We use the triplet loss as the training goal:
{\small
\begin{equation}
\begin{split}
L_{triplet} = \sum_{(\vect{z}_{a}, \vect{z}_{p}, \vect{z}_{n}) \in \mathcal{Z}} max(0,
\vect{d}_a(\vect{z}_{a}, \vect{z}_{p}) - \vect{d}_a(\vect{z}_{a}, \vect{z}_{n}) + \theta),
\end{split}
\end{equation}
}
where $(\vect{z}_{a}, \vect{z}_{p}, \vect{z}_{n})$ denotes an instance of triplet where $\vect{z}_{a}$ is the anchor, $\vect{z}_{p}$ is a candidate from the same trajectory (positive sample), and $\vect{z}_{n}$ is a candidate from a different one (negative example). $\vect{d}_a(\vect{z}_1,\vect{z}_2)$ denotes the Euclidean distance between  $\vect{z}_1$ and $\vect{z}_2$, which is called the \emph{appearance distance}.

We apply a convolutional neural network to learn the embedding $\vect{z}$, with the architecture illustrated in ``Triplet Network'' module of Fig.~\ref{fig:workflow}. The convolutional feature maps of original target images are flattened, fed into the fully connected layers and finally normalized by an L2-Normalization (L2-Norm) layer. The output of L2-Norm is the 128-dimensional \emph{appearance embedding}, $\vect{z}$. 

We adopt online sampling to generate more instances. The sampling strategy includes: 1) we sample $N$ persons and $l$ instances of each per batch;  
2) we divide each trajectory into segments so that the temporal distance within a segment is no longer than $\tau$; 
3) in each batch, we select one segment to draw samples of a person, with target detection boxes randomly shifted around at each frame as data augmentation; 
4) we add in the targets that the detector has missed (but in the labeled ground truth) into the training set.

Similar to FaceNet~\cite{schroff-et-al:2015}, we make full use of every positive sample pair, but only use the most violating $K$ negative samples selected by hard negative mining, together with $K$ randomly drawn violating samples to construct the triplet set $\mathcal{Z}$. Consequently, the process produces a total number of $|\mathcal{Z}| = N\times l\times (l-1) \times 2\times K$ triplets per batch.%

\subsection{Two-level Association}
\label{sec:twolevelassociation}
The difficulty of data association is different in sparse and dense scenes. Inspired by Huang \etal~\cite{huang-et-al:2008}, we use a two-level association paradigm. The method connects neighboring boxes into tracklets at the low level and performs the end-to-end tracklet association at the high level.  
\subsubsection{Tracklet Generation}
\label{sec:tracklet_generation}
Performing low-level association with simple model helps us cut down the number of nodes and candidate edges. To learn a robust low-level affinity model, we take both appearance and spatial features into consideration as the input of the affinity model. Then we use the Hungarian algorithm~\cite{kuhn-munkres:1955} based on the output of the affinity model to decide whether candidate pairs between adjacent frames should be matched. The track formed by the matched boxes, a.k.a. \emph{tracklets}, is used in the second stage association. 

\para{Appearance feature.} We use the embedding distance between  two candidates of the pair as appearance feature, denoted as $\vect{d}_a(\vect{z}_1, \vect{z}_2)$ (from Section~\ref{sec:appearance}). The metric reflects how similar two candidates look alike in visual. 
\label{section:appearance}

\para{Spatial feature. }
For low-level association, we only take the candidate pairs in adjacent frames into consideration. We define the \emph{relative position distance} as
{\small
\begin{equation}
\begin{split}
\vect{d}_{p}(&P_1, P_2)= \\
&(\frac{|P_{x_1} - P_{x_2}|}{\sqrt[]{P_{w_1}P_{w_2}}}, \frac{|P_{y_1} - P_{y_2}|}{\sqrt[]{P_{h_1}P_{h_2}}}, \frac{|P_{w_1} - P_{w_2}|}{P_{w_2}}, \frac{|P_{h_1} - P_{h_2}|}{P_{h_2}}),
\end{split}
\end{equation}
}where $(P_{x_i}, P_{y_i}, P_{w_i}, P_{h_i})_{i=1,2}$denotes point-size coordinates of the two bounding boxes of the candidate pair.  

\para{Feature fusion.  }
Concatenating the appearance distance $\vect{d}_{a}(\vect{z}_1, \vect{z}_2)$, the relative position distance, along with humanity score for the two patches as feature input, we train an MLP classifier to predict the affinity score $y\in [0,1]$ between candidate pairs. A score close to $1$ indicates the pair bounding boxes belong to same trajectory and score close to $0$ otherwise. Assuming there are $r_t$ detections in frame $t$ and $r_{t+1}$ detections in frame $t+1$, we construct a matching matrix $\mathcal{Y}\in R^{r_t\times r_{t+1}}$ from affinity score of the candidate pairs. Then we use the Hungarian algorithm to calculate matching pairs. From a conservative perspective, we only keep matching results with very high confidence at this stage by applying the Eq.(3) of ~\cite{huang-et-al:2008}, and leave uncertain matchings to next association phase where we consider more extended sequence information.      

\subsubsection{Tracklet Associaton}
Given tracklets from Section~\ref{sec:tracklet_generation}, we train an \emph{association model} based on an end-to-end learnable network flow formulation to associate the tracklets into trajectories. 
\label{sec:association}

\para{Problem Formulation.  }
Fig~\ref{fig:workflow} illustrates the structure of our network flow paradigm in ``End-to-end Tracklet Association'' module. We use nodes to represent tracklets, and edges to represent candidate tracklet pairs that may be associated. The \emph{source} and \emph{sink} are two auxiliary nodes to indicate the initialization and termination of trajectories.

 The bi-level optimization problem for MOT is formulated as follows~\cite{schroff-et-al:2015}. We solve for the parameter $\vect{\theta}$ with: 
{\small
 \begin{equation} \label{eq:lp}
    \begin{aligned}
        \argmin_{\vect{\theta}} \mathcal{L}(\vect{x}^{gt}, \vect{x}^\star) = \sum_{\kappa \in \{det, init, term\}} \sum_i w_i(x_i^{\kappa, gt} - x_i^{\kappa, \star})^2
    \end{aligned}
    \end{equation}
    \begin{equation} \label{eq:lower_smooth}
    \begin{split}
   \vect{x}^\star = \argmin_{\vect{x}} \sum_i  c^{det}_i(\vect{f}_u, \vect{\theta}_1) x_i^{det} + \sum_{i,j} c^{link}_{i,j}(\vect{f}_p, \vect{\theta}_2)x^{link}_{i,j} \\
  + \sum_ic^{init}_i x_i^{init} + \sum_i c^{term}_i x_i^{term}
    \end{split}
    \end{equation}
    \begin{subequations} 
        \vspace{-2mm}
    s.t.,
    \vspace{-2mm}
    \begin{equation} \label{constraint:flatten_x}  
   \vect{x} = [\cdots, x^{det}_i, x^{init}_i, x^{term}_i,\cdots,  x^{link}_{i,j}, \cdots]\in \mathcal{R}^{M\times 1}
        \hspace{0mm}
    \end{equation} 
    \begin{equation} \label{constraint:equality}  
       \displaystyle 
         \vect{C}\vect{x} = \vect{0}, 
        \hspace{20mm}
    \end{equation} 
    \begin{equation} \label{constraint:inequality}  
        \displaystyle 
       \vect{A}\vect{x} \le \vect{b}
        \hspace{20mm}
    \end{equation} 
    \end{subequations}
}

where $\vect{x}^{\{det, link, init, term\}} \in \{0,1\}$ represents whether the unary, pairwise, source-to-node and node-to-sink edges are connected, and $\vect{c}^{\{det, link, init, term\}}$ are their corresponding costs. The weights $w$ is a hyper-parameter balancing the importance of each edge. For more details on definition of $\vect{A}, \vect{B}, \vect{C}$, please refer to the work of Schulter~\etal~\cite{schulter-et-al:2017}. The earlier work~\cite{schulter-et-al:2017} uses log-barrier and basis substitution to eliminate constraints, and achieves the expected partial derivative of $\frac{\partial\mathcal{L}}{\partial\vect{c}}$(Eq. (10) of ~\cite{schulter-et-al:2017}). We find the solution to be unnecessary and defective, so we give our solution which can achieve the goal without the complicated matrix multiplication, as stated later in this section.



%

Based on the problem formulation, we make the following three significant improvements:

\para{Improvement 1: Using tracklet-level features.}
\label{sec:tracklet_feature}
We define the tracklet set as $\{F_i, i=1,\cdots,n\}$. We model the unary cost $\vect{c}^{det}$ and pairwise cost $\vect{c}^{link}$ as $\vect{c}^{det}(\vect{f}_u, \vect{\theta}_1)$ and $\vect{c}^{link}(\vect{f}_p, \vect{\theta}_2)$ that are fit by multi-layer perceptrons (MLP). $\vect{f}_u$ is the unary tracklet feature, and $\vect{f}_p$ is the feature extracted from tracklet pairs. $\vect{\theta}_1, \vect{\theta}_2$ are MLP parameters w.r.t. unary and pairwise functions.
 For a pair of connected tracklets $F_i$ and $F_j$, $t_i$ and $t_j$ denote the frame at the tail of $F_i$ and the head of $F_j$, respectively.  Variables $\vec{h}_i$, $\vec{d}_i$, and $\vect{\overline{z}}_i$ are the humanity scores, detection scores, and the average embedding of tracklet $F_i$.  $\mathcal{S}$ and $\mathcal{A}$ denote the area and aspect ratio of the bounding box. Furthermore, we denote the time gap between two connected tracklets $F_i$ and $F_j$ as $\Delta t = |t_i - t_j|$. We use $\hat{P}^{t}_i$ and $\hat{P}^{h}_j$ to denote the forward and backward Kalman filter estimated position for $F_i$ (from $t_i$ to $t_j$) and $F_j$ (from $t_j$ to $t_i$), respectively.  We also include in the absolute position distance between $F_i$ and $F_j$, denoted as $\vect{d}_p(P^t_i, P^h_j)$, in case that Kalman Filter does not work robustly on short tracklets. We divide the position distance terms by $\Delta t$ to support association across a long time period.
With these notations, we can write $\vect{f}_u$ and $\vect{f}_p$ as
{\small
\vspace{-2mm}
	\begin{equation}
	\begin{split}
	& \vect{f}_u(i) = (\vect{median}(\vec{h}_i), \vect{median}(\vec{d}_i), |F_i|)\\
	&\vect{f}_p(i,j) = (\vect{d}_a(i,j), \vect{d}_A(i,j), \vect{d}_S(i,j), \vect{d}_p^U(i,j), |F_i|, |F_j|, \Delta t), \\
	& \text{where} \quad \vect{d}_a = \vect{d}_{a}(\vect{\overline{z}}_i, \vect{\overline{z}}_j), \quad 
	\vect{d}_\mathcal{A} = \frac{\mathcal{A}_i^{t}}{\mathcal{A}_j^{h}}, \quad
	\vect{d}_\mathcal{S} = \frac{\mathcal{S}_i^{t}}{\mathcal{S}_j^{h}}, \\
	&\quad\quad\quad  \vect{d}_p^U = (\frac{\vect{d}_{p}(\hat{P}^{t}_i, P^{h}_j)}{\Delta t}, \frac{\vect{d}_{p}(P^{t}_i, \hat{P}^{h}_j)}{\Delta t}, \frac{\vect{d}_{p}(P^{t}_i, P^{h}_j)}{\Delta t})
	\end{split}
	\end{equation}
	\vspace{-4mm}
}

\para{Improvement 2: Fixing training deficiency with approximate gradient.  }
\label{sec:approximte_gd}
To solve the bi-level optimization problem Eq.~\ref{eq:lp}-~\ref{eq:lower_smooth}, we find it is unnecessary to calculate the partial derivative $\frac{\partial\mathcal{L}}{\partial\vect{c}}$ precisely~\cite{schulter-et-al:2017}, not to mention the defect of the gradient formula they provide. The explaination is three-fold: 
{\small
\begin{equation}	
\label{eq:dL_dc}	
\frac{\partial \mathcal{L}}{\partial \vect{c}} = -t\cdot \vect{B} [\vect{B}^T \frac{\partial^2 P}{\partial {\vect{x}^\star}^2} \vect{B}]^{-1}\vect{B}^T\frac{\partial\mathcal{L}}{\partial \vect{x}^\star} \\
\end{equation}
\begin{equation}
\label{eq:ddp_ddx2}
 \frac{\partial^2{P}}{\partial \vect{x}^2} = Diag(\frac{1}{(1-x^\star_i)^2}+\frac{1}{{x^\star_i}^2}). 
\end{equation}
 }
1) When the temperature $t=10M$, Eq.~\ref{eq:ddp_ddx2} results in an unstable gradient of $\frac{\partial\mathcal{L}}{\partial\vect{c}}$. when $\vect{x}$ approaches $\{0, 1\}$, $\frac{\partial^2{P}}{\partial \vect{x}^2}$ is large, and the corresponding gradient $\frac{\partial \mathcal{L}}{\partial \vect{c}}$ becomes small, as a result the loss descends slowly at the beginning. On the other hand, when $\vect{x}$ approaches 0.5, $\frac{\partial^2{P}}{\partial \vect{x}^2}$ is relative small, $\frac{\partial \mathcal{L}}{\partial \vect{c}}$ increases sharply because of the large temperature. These extreme gradient values harms the training paradigm of deep learning, which usually uses fixed learning rates that are independent to the gradient order of each iteration. It's easily to get stuck or start oscillating during training.  2) Eq.~\ref{eq:dL_dc} does not guarantee the movement smoothness of $\frac{\partial\mathcal{L}}{\partial\vect{x}}$ because $B[B^T\frac{\partial^2{P}}{\partial \vect{x}^2} B]^{-1}B^T \in \mathcal{R}^{M\times M}$ is not invertible. It indicates that even though  $\vect{c}$ is moving towards a dedicated direction temporarily, the impact on $\vect{x}$ is not predictable. 
3) Moreover, When moving to a combination of values of $\vect{c}$ and $\vect{x}$, the original constrained linear programming becomes hard to solve as it takes a long time to jump out of the bad point.

Thus, we doubt the plausibility of the chain rule after performing the basis substitution in Eq.~\ref{eq:lower_smooth}. However, the formulation is still valuable, because we can use an approximate gradient rather than the accurate gradient to solve the above problems. Our idea is quite straightforward but unexpectedly works well: intuitively, if $\frac{\partial \mathcal{L}}{\partial x_i} > 0$,  it means $x_i$ is larger than the expected value $x_i^{gt}$ (in our case, it also indicates that $x_i^{gt} = 0$). Then $x_i$ is supposed to be turned down towards $0$ so that its corresponding edge is not chosen into the final trajectory. To achieve this, the corresponding cost $c_i$ should be enlarged. In contrast, if $\frac{\partial \mathcal{L}}{\partial x_i} < 0$, we should reduce the value of $c_i$ so that $x_i$ is increasing towards $1$. To save computation, we simply set $\frac{\partial L}{\partial \vect{c}} = - \frac{\partial L}{\partial \vect{x}}$.

\begin{figure}[t]
\begin{center}
\begin{subfigure}[h]{0.49\linewidth}
 \includegraphics[width=1.\linewidth]{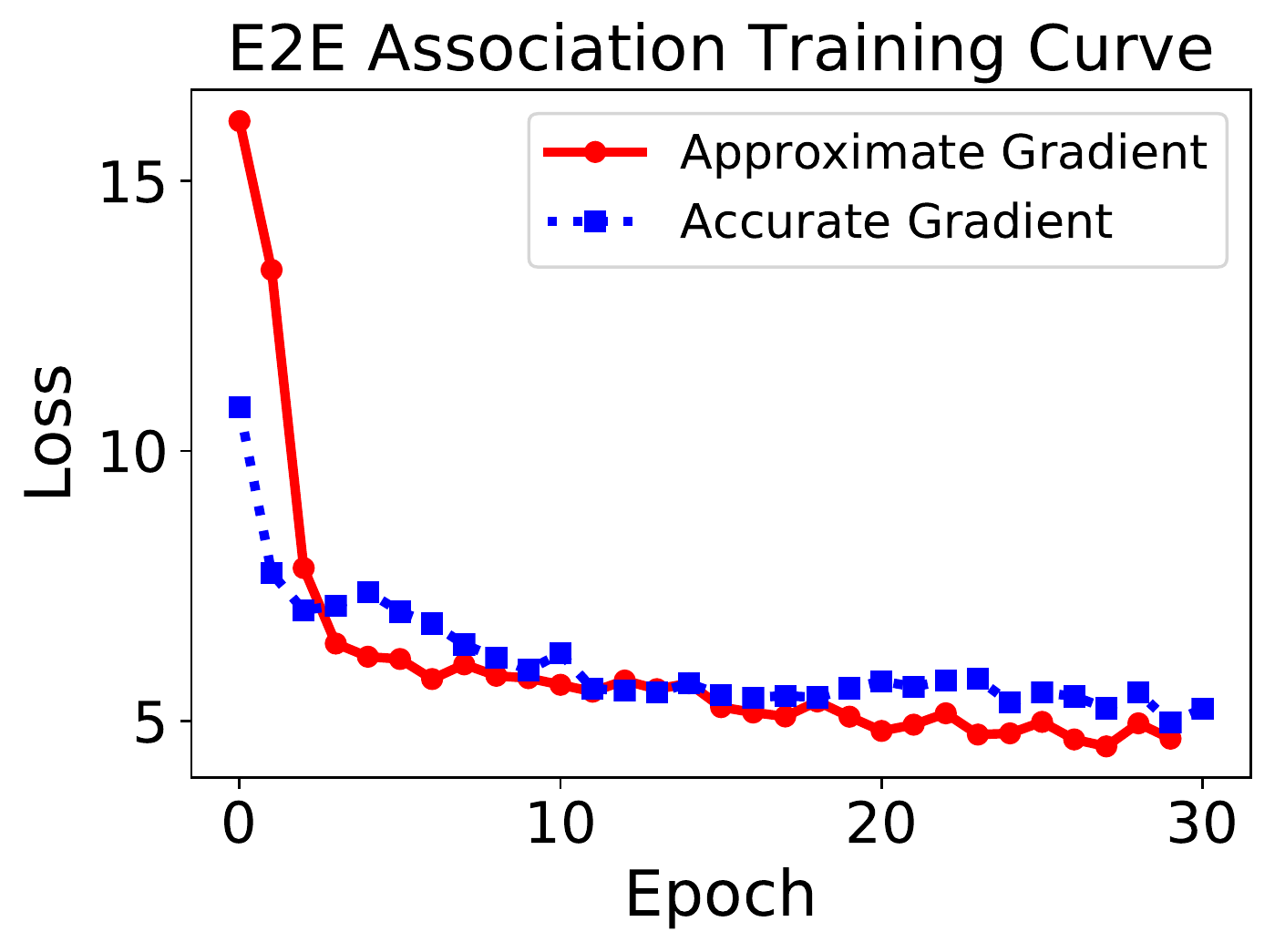}
 \caption{}
 \label{fig:e2e_train_curve}
   \vspace{-2mm}
 \end{subfigure}
 \begin{subfigure}[h]{0.5\linewidth}
  \includegraphics[width=1.\linewidth]{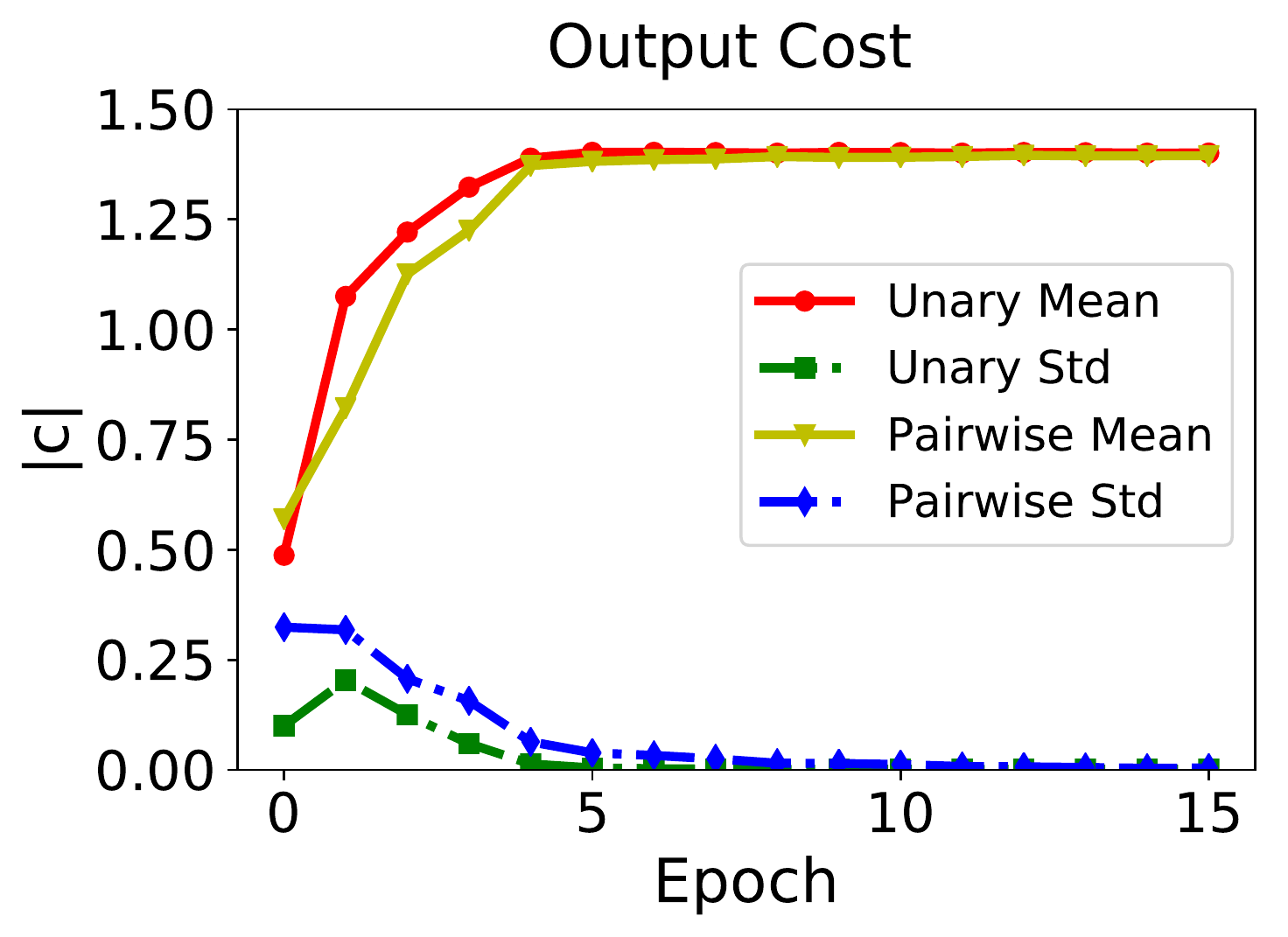}
  \caption{}
  \label{fig:train_cost}
  \vspace{-2mm}
  \end{subfigure}
\end{center}
\vspace{-4mm}
   \caption{(a) The learning curves of the end-to-end association by calculating the accurate gradient or approximate gradient. (b) The transition curve of the unary and pairwise costs in absolute form. }
\label{fig:e2e_train}
\vspace{-3mm}
\end{figure}

\para{Improvement 3: Bound the range of cost parameters.  }
\label{sec:hyper}
 Also, we find Schulter~\etal~\cite{schulter-et-al:2017} do not constrict the value of the cost vector $\vect{c}$. It is risky because we never know what value the cost $\vect{c}$ will converge to. In our experiments, the value of $\vect{c}$ sometimes diverges to infinity if the learning rate is not carefully set. Even converged, the final converge point is not predictable, and the computation of the constrained LP problem with large value is costly. Instead, we bound the value of $\vect{c}^{det, link}$ to range $[-\gamma, \gamma]$ by adding a $tanh$ function to the output of the MLP, and we can arbitrarily initialize $\vect{c}^{init,term}$ to some constant $\beta \in [-\frac{\gamma}{2}, \frac{\gamma}{2}]$. Fig~\ref{fig:train_cost} shows the absolute value of $\vect{c}^{det}$ and $\vect{c}^{link}$ finally converges at the line of $2\beta$, indicating that $c^{det}$ and $c^{link}$ can be automatically adjusted to the boundary of tracklet TP and FP: $-2\beta$, only if the value of $-2\beta$ is reachable.

Fig.~\ref{fig:e2e_train_curve} shows the learning curve by using the approximate gradient. In practice, we find the training speed with our approximation approach is over $20\times$ larger than the original, without any sacrifice of performance. It is because the accurate gradient in Eq.~\ref{eq:dL_dc} contains high dimensional matrix multiplication and inversion, while the approximate one is a lot cheaper.

\para{Other parameter choices. } 
To deal with long video sequences, we use a $W$-frame sliding window to segment video sequences into subgraphs. We can use a larger window and steps because tracklets generates much fewer candidate pairs comparing to detection responses. Tracklets with head or tail locating in the temporal window are included as nodes in the subgraph. We set the stepping size to $\frac{1}{4}\sim \frac{1}{2} W$ to preserve the overlapping necessary for merging subgraph associations into the longer tracks. We use Hungarian matching over intersections between tracklet sets to merge these subgraphs. An example of subgraph merging result shows in Figure~\ref{fig:workflow}. Here, rectangle blocks are tracklets with their widths showing different tracklet lengths, and three subgraphs generate three hypothetic long tracks. As depicted in the figure, \emph{Subgraph 1} produces three intermediate long tracks while \emph{Subgraph 2} produces four. These intermediate tracks generate a matching matrix in $\mathcal{R}^{3\times 4}$, with elements being the number of overlapping blocks between tracks. Then the Hungarian algorithm is used to find the matching with largest overlaps, and merge-sort is used to combine tracklets in order.


The setting of weights $\vect{w}$ in Eq.~\ref{eq:lp} for tracklet fusion is more straightforward than those at the detection response level. Existing work~\cite{schulter-et-al:2017} empirically sets $w$ lower for a punishment of ambiguous links like FP-FP(two false positives) and TP-TP+Far(true positives with the same identity but distant). However, in the case of tracklets, the weighting directly reflects on the final metric MOTA.  For instance, a FP edge between tracklets leads to $|F_i^{tail} - F_j^{head}|$ FP boxes in the final trajectory; in contrast, a FN pair losses $|F_i^{tail} - F_j^{head}|$ expected boxes in the final trajectory, and same deduction applies to unary terms. Thus, we set the weights of $x^{init}_i, x^{term}_i$ to 1, the weights of $x_i^{det}$ to the length of tracklets, and the weights of $x^{link}_{i,j}$ to the time gap between tracklet pairs. Our experiments in Section~\ref{sec:experiment} show that the weighting strategy improves performance.

\subsection{Post Processing}
There are certain detection gaps between tracklets caused by occlusions, and we fill them in post-processing.  Bilinear interpolation is a general approach, but it is prone to introduce errors.  
To validate the legitimacy of the interpolated boxes (a.k.a virtual patches), we feed the corresponding virtual patches into Proposal Alignment Network to get the humanity and regressed boxes, and compute appearance distance corresponding to the last box in the filled trajectory. 
We discard the interpolated boxes with low humanity or appearing very different from others in trajectory.

\section{Evaluation}
\label{sec:experiment}
\subsection{Implementation Details}
\para{Datasets.} We evaluate our approach on both the 2D MOT2016 and MOT2017 Benchmarks. MOT2016 offers a variety of 14 video sequences (7 for training and 7 for testing)~\cite{milan-et-al2:2016} which are captured by static or moving cameras indoors and outdoors. MOT2016 provides with the detection responses of DPM detector for training and testing. MOT2017 contains the same sequences but provides with more precise ground-truth annotations. The organizers release the detection results of Faster R-CNN and SDP in addition to that of DPM so that researchers are supposed to submit tracking results under the three detectors. 

\para{Platform.} All of our experiments are conducted on a 1.2GHZ Intel Xeon server with 8 NVIDIA TITAN X GPUs. The deep learning framework we use is MXNet~\cite{chen-et-al:2015}.

\para{Appearance model.} We use an ImageNet pre-trained ResNet-50~\cite{he-et-al:2016} as the CNN backbone for both the Proposal Aligner and the Triplet Network. We set the input size to $128\times 64$, and use ReLU for activation and ADAM~\cite{kingma-ba:2014} optimizer to train the networks.  The base learning rate is set to $1e-4$, $1e-6$, respectively. The hyper-parameters of sampling are set as: $l=8, N=50, K=100, \alpha=0.8, \tau=60$.

 
%
\para{Bi-level optimization training.} We use a three-layer MLP with Leaky ReLU activation for the pairwise network. The reason of using Leaky ReLU rather than ReLU is that the regression target is near a single value: both $\vect{c}^{det}$ and $\vect{c}^{link}$ are expected to converge at around of $-2\beta$ so that the neural work tends to move simultaneously at same pace for all samples, and using ReLU may wipe out sample difference at very beginning.  We choose \textit{cvxpy}~\cite{diamond-boyd:2016} as the convex problem solver, and also ADAM~\cite{kingma-ba:2014} as the MLP optimizer. We set the value of $\beta$ to $0.7$ and $\gamma$ to $5$ in all experiments.
%

\para{Metrics. }  For evaluation, we report the following metrics: 1) the most commonly used metric, $MOTA = 1 - \frac{FP + FN + IDS}{GT}$, which provides a combination of the false positive (FP), false negative (FN) and ID switch (IDS) frequency among all trajectories against the ground truth (GT); 2) mostly tracked (MT) and 3) mostly lost (ML) that provide an indication of the trajectory fragmentation.

\subsection{Comparison to the State-of-the-art}
\label{sec:state-of-art}
Table~\ref{tab:resultmot2016} shows our MOT2016 benchmark results in comparison with the current competitive approaches. We observe that 1) our MOTA of $49.0$ outperforms all published results as of March of 2018. We believe this high MOTA is a direct result of low FN of $82,506$, attributing to both the alignment and association approach. 

%

Table~\ref{tab:resultmot2017} shows our result on MOT2017 benchmark. We haven't performed any fine-tuning on this dataset based on MOT2016, and 3 detectors share the same set of configuration. Our result ranks the 3rd place with MOTA of 51.5 by April 9th, 2018, with the 1st and 2nd place being non-published. The performance on Faster R-CNN is the highest on board. By fine-tuning the parameters, it's possible to achieve higher performance. We choose not to do so but only show the fact that our method applies to any detectors, and better detector benefits tracking result.

{
 \begin{table}
 \fontsize{5}{5}\selectfont
 \footnotesize
   \vspace{-4mm}
 \begin{center}
   \begin{minipage}[t]{1.0\linewidth} 
 \begin{tabular}{|l|l|l|l|l|l|}
\hline
Tracker &  MOTA$\uparrow$ & MOTP$\uparrow$ & IDS$\downarrow$ & MT$\uparrow$ & ML$\downarrow$ \\
\hline
$\text{GMMCP}^{~\cite{dehghan-et-al:2015}}$& 38.1 & 75.8	 & 937 & 8.6 & 50.9 \\
$\text{QCNN}^{~\cite{son-et-al:2017}}$ & 44.1 & 76.4 & 745 & 14.6 & 44.9 \\
$\text{MHT\_DAM}^{~\cite{kim-et-al:2015}}$ & 45.8 & 76.3 & 590 & 16.2 & 43.2 \\
$\text{NOMT}^{~\cite{choi:2015}}$& 46.4 & 76.6 & \textbf{359} & 18.3 & 41.4 \\
$\text{AMIR}^{~\cite{sadeghian-et-al:2017}\footnote{Online tracker}}$ & 47.2 & 75.8 & 774 & 14.0 & 41.6 \\
$\text{FWT}^{~\cite{henschel-et-al:2017}}$ & 47.8 & 75.5 & 852 & 19.1 &  38.2 \\ 
$\text{LMP}^{~\cite{tang-et-al:2017}}$ & 48.8 & \textbf{79.0} & 481 & 18.2 & 40.1 \\
\hline
\sysname & \textbf{49.0} & 78.0 & 899 & \textbf{19.1} & \textbf{35.7} \\
\hline
 \end{tabular}

    \end{minipage}
     \vspace{-2mm}
       \caption{Results on the MOT2016 test data.}
  \label{tab:resultmot2016}
 \vspace{-5mm}
 \end{center}

 \end{table}
}
{\small
\begin{table}[t]
\centering
\vspace{2mm}
  \begin{minipage}[t]{1.0\linewidth} 
 \begin{tabular}{|l|l|l|l|l|l|} 	
\hline
 Tracker & Detector & MOTA$\uparrow$ &  IDS$\downarrow$ & MT$\uparrow$ & ML$\downarrow$ \\
\hline
  \multirow{ 4}{*}{\cite{kim-et-al:2015}}  & DPM & 44.6 & 593 & 15.3 & 45.2 \\
& FRCNN & 46.9  & 742 & 18.6 & 34.7 \\
& SDP & 60.6  & 979 & 28.7 & 30.8 \\
& Total & 50.7 & 2,314 & 20.8 & 36.9 \\ 
\hline
\multirow{ 4}{*}{\cite{henschel-et-al:2017}} & DPM & 46.4 & 833 & 18.2 & 40.8 \\
& FRCNN & 48.2 & 780 & 18.5 & 35.9 \\
& SDP & 59.4 & 1035 & 27.6 & 29.0 \\
& Total & 51.3& 2,648 & 21.4 & 35.2 \\
 \hline
 & 
DPM & 45.9 & 552 & 18.3 & 44.9 \\
eHAF& FRCNN & 47.8 & 605 & 20.0 & 36.7 \\
17\footnote{Anonymous submission
}& ${\star}$SDP & 61.8 & 677 & 31.9 & 32.1 \\
& Total & 51.8 & \textbf{1,834} & \textbf{23.4} & 37.9 \\
 \hline 
& ${\star}$DPM & 49.8 &  1,209 & 19.5 & 36.2 \\
CNN & FRCNN & 46.5 & 828 & 17.6 & 34.4 \\
{\_search\footnote{Non-published work from Hikvision Research Institute}}& SDP & 60.2 &  1022 & 27.1 & 29.4 \\
& Total & \textbf{52.2} & 3,059 & 21.4 & \textbf{33.3} \\
 \hline

 \multirow{ 4}{*}{\sysname}&  DPM & 47.0 & 698 & 15.2 & 42.3 \\
& ${\star}$FRCNN & 48.3 & 866 & 17.5 & 36.6 \\
& SDP & 59.2 & 1,029 & 29.2 & 27.6 \\
& Total & 51.5 & 2,593 & 20.6 & 35.5 \\
\hline
    \end{tabular}

   \end{minipage}
           \caption{Result on the MOT2017 test set. The $\star$ denotes the best performance achieved yet of each detector.}
       \label{tab:resultmot2017}
      \vspace{-4mm}
 \end{table}
}

\subsection{Ablation Study}

To give a transparent demonstration of the impact of each component we have introduced, we perform comparative experiments at each stage on MOT2016. For all experiments in this section, we use the last two training sequences (MOT16-11, MOT16-13) for validation and the rest for training. 

\para{Proposal aligner \& proposal selector. } We evaluate the effects of the proposal aligner and proposal selector, which have shown significant improvement on the final result. We denote the Aligner as \emph{AL}, non-maximum suppression and the score filtering of the selector as \emph{S-NMS} and \emph{S-SF}.
We compute the IOU ratio against the ground truth. We define an IOU greater than 0.5 as a true positive (TP), and otherwise as a false positive (FP). Under this setup, we get $8,500$ TPs, $5,445$ FPs from the two validation sequences.  We separately evaluate these elements, and use the best association setting of \sysname to do the two-level association. Table ~\ref{tab:abppr} summarizes the results.

We notice that by performing the proposal alignment, we can rectify $90$ FPs to TPs in the validation set. It indicates that tighter boxes can not only benefit feature extraction but also explicitly enhance the recall of proposals. Besides, by performing S-NMS and S-SF, the number of FP decreases from $5355$ to $1244$, with a sacrifice of only $283$ TPs. The MOTA after alignment and selection is 13.9\% larger than that of raw input. To clarify, we use 0.7 as NMS threshold, and remove the boxes with humanity lower than 0.1 and the detection score lower than 0 in this experiment. Moreover,  the experiment shows MOTA decreases slightly in MOT16-13. It is because the sequence contains many people in shadow, which is a rare scenario in the training set. Thus they are assigned low humanity, and S-SF filters them out. However, the case happens rarely in general circumstances, and we employ all these strategies in other experiments.

To validate whether all detectors benefit from the aligner and selector, we report the statistics on the MOT2017 dataset in Table~\ref{tab:pre2017}. From training set, we see remarkable FP decrease on DPM and respectively a high improvement on MOTA. While for Faster R-CNN and SDP which have coordinate regression module in their algorithms, the benefit is less but still visible. It's partially because for these detectors the alignment is good enough.

{\small    

 \begin{table}
\centering
   \begin{minipage}[t]{1.0\linewidth} 

\begin{tabular}{|l|l|l|l|l|l|l|l|}
\hline
 Case & \multicolumn{3}{c}{MOTA$\uparrow$} & RTP$\uparrow$ & RFP$\downarrow$ \\
\hline
 Seq & 11+13 & 11 & 13 & 11+13  & 11+13    \\
\hline
 DEFAULT & 32.5 & 49.9 & 18.6 & 8,500 & 5,445 \\
 AL & 34.5 & 51.2 & 21.2& 8,590 & 5,355  \\
 AL + S-NMS & \textbf{37.3} & 53.8& \textbf{24.1} & 8,512 & 3,181 \\
{AL + S-NMS } &   \multirow{ 2}{*}{37.0} &  \multirow{ 2}{*}{\textbf{54.9}} &   \multirow{ 2}{*}{22.6} &   \multirow{ 2}{*}{8,307} &   \multirow{ 2}{*}{1,244}\\
    + S-SF  &&&&&\\
\hline
 \end{tabular}
   \caption{Performance of the proposal aligner and the proposal selector on the validation set. RFP and RTP denote the residual FP and TP boxes after each step.}
  \label{tab:abppr}
 \end{minipage}
 \end{table}
 }

 {\small
 \begin{table}
\centering
  \begin{minipage}[t]{1.0\linewidth} 
 \begin{tabular}{|l|l|l|l|l|l|}
\hline
 Detector &  RTP/OTP & RFP/OFP & MOTA(R/O)\\
\hline
 DPM & 50,160/52,760 &  2,890/27,030 & 34.9/47.0  \\
 FRCNN & 64,235/64,252 & 3,316/3,387 & 46.0/48.3\\
 SDP & 78,349/78,623  & 3,733/4,164 & 56.2/59.2  \\
\hline
 \end{tabular}
   \caption{Effect of proposal aligner/selector under different detectors on the MOT2017 dataset. We report the TP/FP on MOT2017 training set as a reference. OTP/OFP denotes original TP/FP; RTP/RFP denotes the remaining TP/FP after performing alignment and selection. The MOTA shown here is evaluated on the test set.}
  \label{tab:pre2017}
 \end{minipage}
    \vspace{-3mm}
 \end{table}
 }

\para{Using tracklets vs. detection responses directly.  } To demonstrate the improvements that tracklets bring, we train an end-to-end association model directly on features extracted from the detection responses.  The feature extraction is the same as in Section~\ref{sec:tracklet_generation}.  We set $W$ to 5 frames and step size to 1 to accommodate the large candidate pairs and ensure overlapping. The two experiments achieve same MOTA at 35.9, while TAT runs much faster than the other. Moreover, TAT achieves best performance 37.9 at $W = 30$ as shown in  Fig~\ref{fig:compare_res}.  In the same configuration, we cannot even obtain a model on detection response level due to the large number of search space.

\para{End-to-end Learned vs. Hand-crafted Affinity.  } The significant improvement of our association method is to incorporate learned-features and cost parameters.  To evaluate the improvements, we compare the following two association methods with \sysname.

\textit{[NETFLOW]}  As a comparative method, we use an independent affinity model together with a standalone inference method to replace the bi-level optimization association paradigm. The proposal alignment, selection and tracklet generation are performed as the same.  We train a 2-class MLP classifier as the affinity model, with label 1 representing tracklets of same target and label 0 representing tracklets of different targets. We denote the output of the MLP as $s^p$, and manually design the edge cost as $c^{link}_i = - log(s^p_i)$, the node cost as a product of 1 minus humanity, i.e. $c^{det}_i = log(\prod_k (1-h_{i,k})), k = 1, \cdots, |F_i|$.  Then we apply the Algorithm 1~\cite{zhang-et-al:2008} as tracklet association approach.

\textit{[E2EP]} It is a transition method between TAT and \textit{[NETFLOW]}. It also uses an end-to-end model, except that we use the same unary feature $c^{det}_i$ as \textit{[NETFLOW]}. However, to avoid applying grid search for $\vect{c}^{init, term}$, we use a linear model to learn the direction and bias. Thus we modify $c^{det} = a*log(\prod_k (1-h_{i,k})) + b, a,b\in \mathcal{R}$, and set $c^{init,term}$ to 0.7.

\textit{[\sysname]} We use both learnable unary and pairwise terms for \sysname, with features defined in Section~\ref{sec:association}.  We construct the unary and pairwise MLP with [8, 4, 1] and [256, 256, 1] network architecture, respectively. 

For each method, we conduct experiments using window sizes ranging from 10 to 100 frames.  Fig~\ref{fig:compare_res} shows the results on the validation set.  Our key observations are:

1) Both \sysname and \emph{[E2EP]} outperform \emph{[NETFLOW]}.  This improvement shows the effectiveness of the end-to-end training.

2) When the window size is small, \sysname significantly outperforms the other two methods, thanks to the automatically tuned cost parameters. 

3) When the window size is over 30 frames, both TAT and \textit{[NETFLOW]} show a significant drop in performance, except for \textit{[E2EP]}. The drop is because performing long-term association has little benefit when 95\% expected links are less than 30 frames. In contrast, it brings in a risk of obscuring the pairwise features because of $\Delta t$. Thus, the false connections will result in more false positives interpolated. However, \textit{E2EP} is robust to the large window size. It is because the hand-crafted feature $\vect{c}^{det}$ has a clear intrinsic relation to affinity, which makes the association results more stable when $\vect{c}^{link}$ becomes inferior.




\begin{figure}[t]
\begin{center}
\begin{subfigure}[h]{0.3\textwidth}
 \includegraphics[width=1.\linewidth]{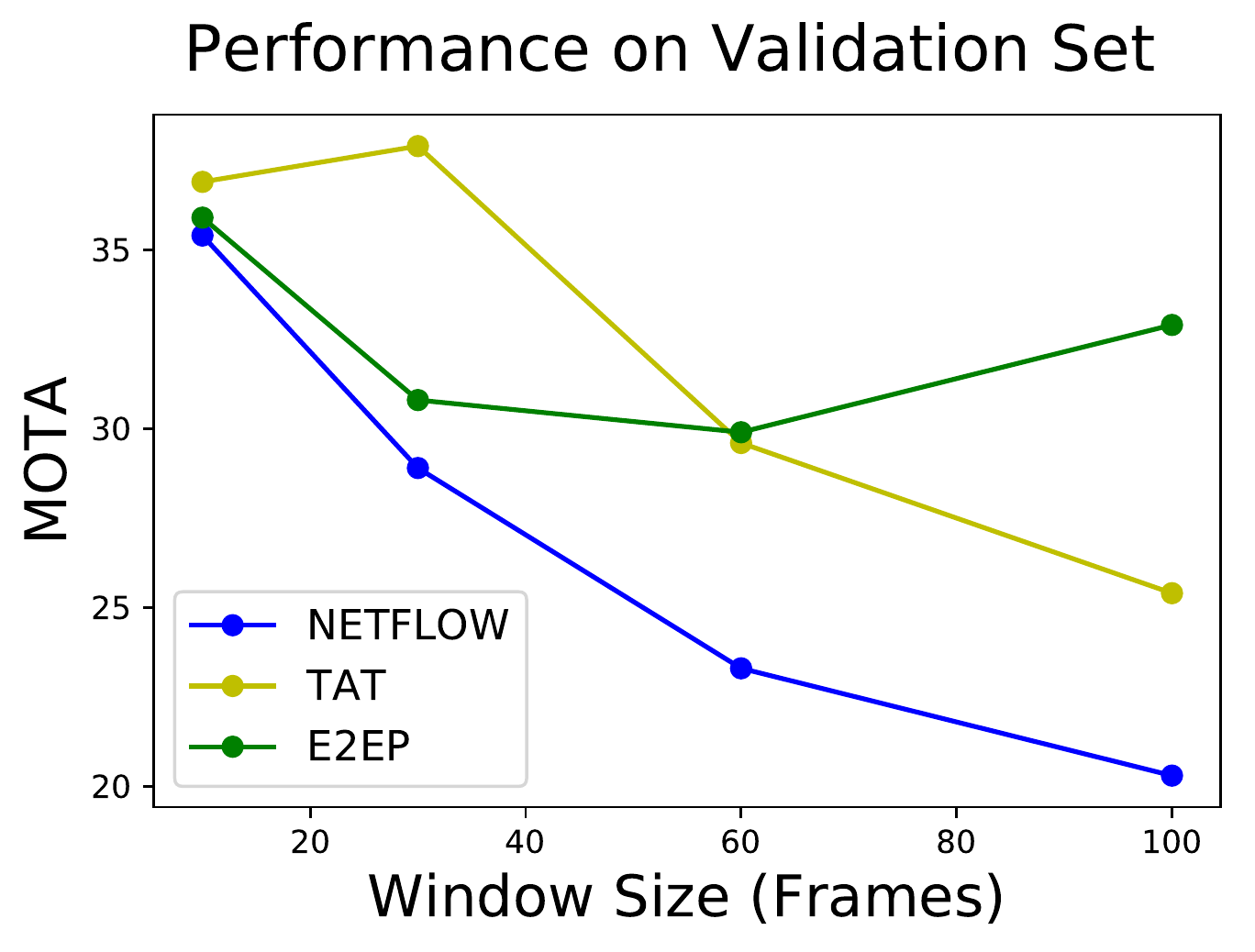}
 \label{fig:compare_res}
 \end{subfigure}
\end{center}
\vspace{-7mm}
   \caption{End-to-end Learned vs. Hand-crafted Affinity. The curve in yellow, green, and blue presenting the MOTA achieved on the validation set at different window length by \emph{TAT}, \textit{[E2EP]}, and \textit{[NETFLOW]}, respectively.}
\label{fig:compare_res}
\vspace{-1mm}
\end{figure}

\para{Weighting. } We claim that tracklet weight $\vect{w}$ in Equation~\ref{eq:lp} should be related to the tracklet length (\emph{TL}) and time gap of connections (\emph{TG}), whose errors directly reflect on the MOTA. From Table ~\ref{tab:weighting} we notice the major improvement comes from TL weighting, which matches our expectation that it costs more to make mistakes on longer tracklets.

 {\small
   \begin{table}
\centering
  \begin{minipage}[t]{1.0\linewidth} 
\begin{tabular}{|l|l|l|l|l|}
\hline
 Case & MOTA$\uparrow$ & IDS$\downarrow$ & FN$\downarrow$/FP$\downarrow$ & MT$\uparrow$/ML$\downarrow$ \\
\hline
 Uniform & 35.9 & \textbf{69} & 12770/\textbf{378} & 11.9/54.6\\
 TL  & 36.9 & 77 & 12550/393 & 11.9/54.6 \\ 
 TG  & 35.9 & 91 & 12716/413 & 11.9/54.0 \\ 
 TL + TG & \textbf{37.0} & 75 & \textbf{12528}/388 & 11.9/\textbf{51.7}\\
\hline
 \vspace{-5mm}
 \end{tabular}
 \caption{The performance on the validation set of variant weighting strategy. The window size is set to 10.}
  \label{tab:weighting}
   \vspace{-2mm}
 \end{minipage} 
 \end{table}
}

\section{Conclusion and Future Work}
\label{sec:conclusion}

Unlike many other tasks in computer vision that adopts end-to-end training, MOT still requires much hand-tuning and optimizations in various stages. \sysname brings MOT an extra step closer to fully end-to-end. With \sysname, we combine the classic tracklet-based association into the new bi-level optimization framework.  
It is easy to integrate additional features since they can be jointly considered in the end-to-end learning framework, and the training process converges much more stable and faster by using the approximated gradient.


Combining tracklets and end-to-end training opens many opportunities for future improvements.   To begin with, we can encode human interactions in the tracklet feature.  Furthermore, we can adopt LSTM-based tracklet features, so that the tracker can perform end-to-end learning from raw data. Lastly, we can investigate how to improve linking consistency around short tracklets in a learning-based association framework.




{\small
\bibliographystyle{ieee}
\bibliography{egpaper_final}
}

\end{document}